\documentclass[12pt]{article}

\usepackage[margin=1in]{geometry}
\usepackage[square]{natbib} 
\usepackage{graphicx}
\usepackage{comment}
\graphicspath{ {./figures/} }
\usepackage{caption}
\usepackage{xcolor}
\usepackage{color}
\usepackage{xspace}
\usepackage{subcaption}
\usepackage{amsmath} 
\usepackage{amssymb}
\usepackage[square]{natbib} 
\usepackage{mathtools}

\DeclareMathOperator*{\argmax}{argmax}

\newif\ifcomments 
\commentsfalse
\commentstrue 

\usepackage{ulem} 

\usepackage{authblk}

\title{Objective discovery of 
dominant dynamical processes with intelligible machine learning}

\author[1]{Bryan E. Kaiser}
\author[1]{Juan A. Saenz}
\author[2,3,4]{Maike Sonnewald}
\author[5]{Daniel Livescu}
\affil[1]{\small{Los Alamos National Laboratory, X-Computational Physics Division XCP-4,}\newline \small{ Los Alamos, NM 87545, USA}}
\affil[2]{\small{Princeton University, Program in Atmospheric and Oceanic Sciences,Princeton, NJ 08540, USA}}
\affil[3]{\small{NOAA/OAR Geophysical Fluid Dynamics Laboratory, Ocean and Cryosphere Division, Princeton, NJ 08540, USA}}
\affil[4]{\small{University of Washington, School of Oceanography, Seattle, WA, USA}}
\affil[5]{\small{Los Alamos National Laboratory, Computer Computational and Statistical Physics Division CCS-2, Los Alamos, NM 87545, USA}}
\date{}
\begin{document}
\maketitle
\begin{center}
{{LAUR-21-25813} \\{e-mail: bkaiser@lanl.gov }}
\end{center}

\begin{abstract}
The advent of big data has vast potential for discovery in natural phenomena ranging from climate science to medicine, but overwhelming complexity stymies insight. Existing theory is often not able to succinctly describe salient phenomena, and progress has largely relied on \textit{ad hoc} definitions of dynamical regimes to guide and focus exploration. We present a formal definition in which the identification of dynamical regimes is formulated as an optimization problem, and we propose an intelligible objective function. Furthermore, we propose an unsupervised learning framework which eliminates the need for \textit{a priori} knowledge and ad hoc definitions; instead, the user need only choose appropriate clustering and dimensionality reduction algorithms, and this choice can be guided using our proposed objective function. We illustrate its applicability with example problems drawn from ocean dynamics, tumor angiogenesis, and turbulent boundary layers. Our method is a step towards unbiased data exploration that allows serendipitous discovery within dynamical systems, with the potential to propel the physical sciences forward.
\end{abstract}

\newpage
\section{Introduction}
A
salient 
class of problems in
the natural sciences is 
nonlinear continuum dynamics problems, 
specifically those pertaining to dynamical systems with uncomputably large numbers of 
degrees of freedom.
Examples span most fields of science; to highlight this diversity we can mention nonlinear waves, plasma dynamics,
earthquake dynamics, general relativity, quantum field theory, 
biochemical reaction-diffusion dynamics, 
fibrillation dynamics, epilepsy, 
and turbulent flows (\cite{strogatz2018nonlinear}), 
as well as fiber optics, droplet formation, 
wrinkling, biofilm dynamics (\cite{callaham2021learning}), 
weather (\cite{vallis2017atmospheric}), and 
climate dynamics (\cite{peixoto1992physics}). 
The extreme sensitivity of such systems 
to infinitesimal 
perturbations (\cite{poincare1905science}), known to popular culture as the butterfly effect 
(\cite{smagorinsky1969problems}), generally prohibits the direct deterministic 
computation of solutions to governing equations because extreme amounts of high-quality data collection and/or computing power are required to marginally increase limits of predictability.

\begin{figure}[h!]
 \centering
 \includegraphics[width=0.7\textwidth]{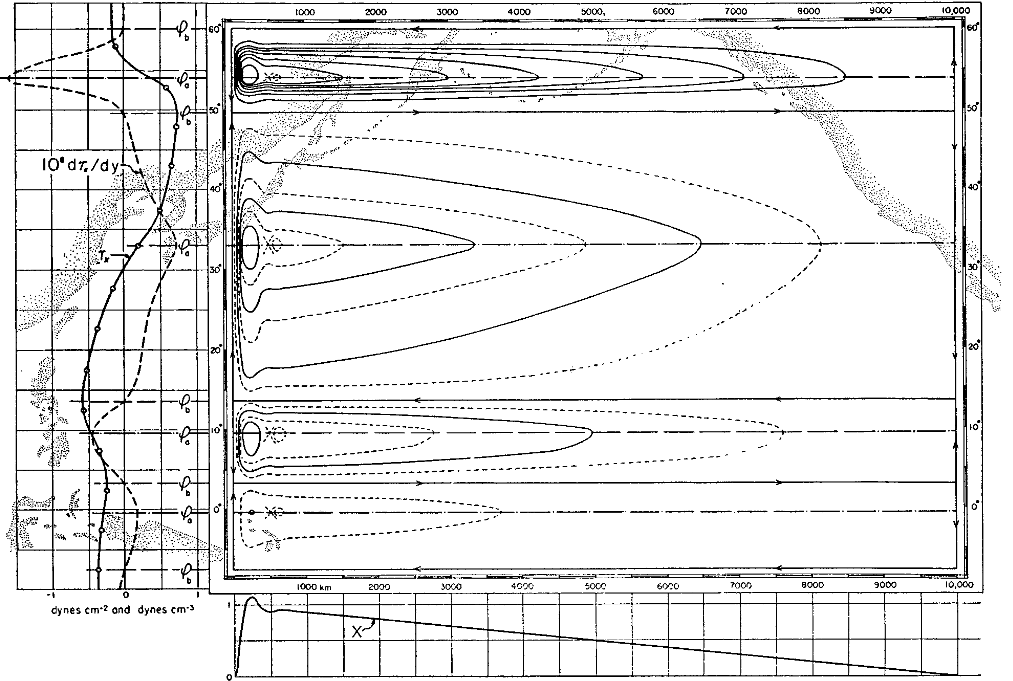}
 \caption{A plot from \cite{munk1950wind} showing 
 the theoretical Pacific Ocean circulation superimposed over the actual coastlines. 
 The interior flow is a dominant balance, 
 referred to as Sverdrup balance, which is 
 adjacent to  
 a western boundary current dominant balance. This idealized circulation correctly 
 hypothesized the dominant mesoscale upper ocean dynamics at mid-latitudes.}
 \label{fig:munk_1950}
 \end{figure}

In lieu of direct computation, 
scientists and engineers have developed a number of alternatives.
A popular method
involves the prediction of 
properties of 
the probability distributions of the relevant variables. The simplest form of this 
strategy is to predict the  
ensemble, spatial, and/or temporal mean of relevant variables.
The dynamics 
of mean variables are sometimes approximately reducible, meaning 
that 
local causality  
can 
be approximately described by 
a \textit{dominant balance} of
selected terms from the governing equation (e.g. \cite{callaham2021learning}). These local 
dominant balances, and the boundaries of the 
solution space in which they 
are a justifiable  
truncation of the full dynamics, define what 
we often refer to as \textit{dynamical regimes}.
In most practical applications the 
dominant balance of a given dynamical regime is 
\textit{non-asymptotic}, meaning that there is no 
parameter that permits a formal expansion of the equation terms which guarantees the truncation error vanishes according to the limits of said parameter. 
Much scientific progress has been made 
by identifying non-asymptotic dominant balances in 
the aforementioned fields. 
For example, in a seminal paper on the theory of mid-latitude ocean circulation, \cite{munk1950wind}
neglected nonlinear governing equation terms to 
identify two asymptotic dominant balances within 
the remaining governing equation terms (Figure \ref{fig:munk_1950}), which means that this model is non-asymptotic in the context of the full governing equation. Indeed, \cite{sonnewald2019unsupervised} identified the same interior dominant balance as \cite{munk1950wind} by using a non-asymptotic unsupervised learning method.

Two key attributes of the concept of a 
dynamical 
regime are that a) it is local and only applicable to 
a certain region of sample space, and b) the dominant equation terms are dominant \textit{relative to the amplitude of the negligible terms in the same region of sample space}.
The first attribute defines a partitioning problem: where 
does one draw the boundary between one regime and 
another, particularly for non-asymptotic continuum dynamics in which 
equation term amplitudes may vary smoothly? The second attribute reveals why simple amplitude thresholds cannot be globally applied to a data set that may exhibit two 
or more dynamical regimes. Instead, the two attributes are closely related and the partitioning of the sample space affects the subset of dominant terms in it, and vice versa. 
Perhaps the most famous example of this pitfall 
is the zero drag paradox of \cite{d1752essai},  
a false conclusion that was inferred 
as a consequence of 
a global threshold that deemed frictional terms negligible everywhere in fluid flow
over a sphere. The paradox wasn't resolved until \cite{prandtl1904flussigkeitsbewegung} 
discovered friction-induced dynamical regimes within a thin boundary layer over the sphere's surface, thus enabling the development of boundary layer theory and, subsequently, modern aircraft design, among other applications. 
We propose that the \textit{dynamical 
regime identification problem}
can be stated as the question: 
\begin{quote}
what is the optimal partitioning of 
the data, such that the dominant terms within 
each partition dominate the negligible terms
by the largest amplitude?
\end{quote}

Until now, the dynamical regime identification problem has not been formulated 
as a mathematical expression. 
In 
this Article, we 1) mathematically formalize the dynamical regimes 
problem as an optimization problem, 2) propose an objective function to quantify the optimal solution to this problem, 3) propose 
a custom dimensionality reduction algorithm for the dynamical regime problem, and 4) propose an unsupervised learning framework (\cite{kohavi1997wrappers}, \cite{dy2004feature}) for solving the problem, drawing 
upon recent successes of unsupervised learning as a tool for partitioning dynamical regimes (\cite{sonnewald2019unsupervised}, \cite{callaham2021learning}) and upon the success of 
dimensionality reduction algorithms (\cite{van2009dimensionality}) for labeling 
dominant balances (\cite{callaham2021learning}). Crucially, the proposed objective function 
defines the optimal for a given clustering algorithm, 
\textit{and} it permits objective comparison of the results of different choices of algorithms. By defining the optimal for the dynamical regime problem, we 
enable the discovery of 
dominant balances without any prerequisite domain knowledge: the 
user need not know what dynamical regimes 
exist in the data before applying our framework.

Throughout this article, we will refer to the dimensionality 
reduction algorithms as \textit{hypothesis 
selection} to emphasize that our unsupervised learning framework is 
analogous to the scientific method: it proposes dominant balance 
hypotheses and subsequently 
tests the fit of those hypotheses to the data. 
This choice of words reflects our perspective that machine learning 
for scientific applications should be developed towards the goal of 
creating an artificial scientific collaborator by systematizing the scientific method, 
rather than continuing the current 
paradigm in which the logic of highly predictive algorithms is extremely difficult to discern (\cite{rudin2019stop}). 
Furthermore, this choice of words also reflects that our 
formal mathematical definition of the problem of  dynamical regime identification
is intentionally constructed so that 
it can be tackled by physical, mathematical, or AI scientists.

\section{Methods}
In this section we propose a formal mathematical definition of the 
dynamical regimes problem, an objective function that defines 
optimal solutions to the dynamical regimes problem, and an unsupervised 
learning framework that solves the dynamical regimes problem. 
\subsection{Problem formulation}
Given the array of data $\mathbf{E} =[\mathbf{e}_1,...,\mathbf{e}_N]$,
consisting of $N$ observations of the $D-$dimensioned vector 
of equation terms $\mathbf{e}_n$, 
we seek to label each observation 
with a $D-$dimensioned hypothesis vector $\mathbf{h}_{n}$,
where $h_{ni}\in\{0,1\}$ for each $n^{th}$ observation of the $i^{th}$ equation term. 
We assume that the equation is closed, $\sum_{i=1}^De_{ni}=0$, 
for all observations. 
The entire array of data 
is labeled by $\mathbf{H}=[\mathbf{h}_1,...,\mathbf{h}_N]$,
and the 
zeros of each hypothesis vector $\mathbf{h}_n$ indicate the equation terms in $\mathbf{e}_n$
that are neglected.
We choose a global objective function 
$\mathcal{M}(\mathbf{E},\mathbf{H})$, such that 
the 
optimal fit of hypotheses for the entire 
data array, $\mathbf{H}_{\text{opt}}$, can 
be obtained by varying the hypotheses $\mathbf{H}$ to 
find $\max\mathcal{M}(\mathbf{E},\mathbf{H})$,
\begin{equation}
\mathbf{H}_{\text{opt}}=
    \begin{cases} 
        \argmax\limits_{\mathbf{H}} \mathcal{M}(\mathbf{E},\mathbf{H})
        \quad : \quad \max \mathcal{M}(\mathbf{E},\mathbf{H})>
        \mathcal{M}(\mathbf{E},\mathbf{1})\\
        \mathbf{1} \hspace{29mm} \quad : \quad  \max \mathcal{M}(\mathbf{E},\mathbf{H})\leq
        \mathcal{M}(\mathbf{E},\mathbf{1})
    \end{cases}\hspace{-1mm},\label{eq:problem_formulation}
\end{equation}
where 
$\mathbf{H}=\mathbf{1}$ is an array of ones indicating 
all equation terms are 
retained for the entire data array. 
We use the notation convention of \cite{bishop2006pattern} in which scalars are presented in italics, lower case bold variables represent one dimensional arrays, and upper case bold variables represent two or higher dimensional arrays. 

We propose Equation \ref{eq:problem_formulation} as 
the definition of the dynamical regime problem, in 
which one seeks to partition the observations $\mathbf{E}$
into different regimes with different 
dominant balances, as labeled 
by $\mathbf{H}_\text{opt}$. The dominant balances within $\mathbf{H}_\text{opt}$ 
can be assigned by traditional, heuristic methods 
for eliminating equation terms, such as using 
characteristic scales to estimate term magnitudes 
(\cite{tennekes1972first}), or they can 
be assigned by partitioning the data using 
clustering algorithms and subsequently 
labeling all observations in each 
cluster with the same dominant balance by 
using dimensionality reduction algorithms 
(\cite{callaham2021learning}). 
Next, we propose a simple, intelligible objective 
function for $\mathcal{M}(\mathbf{E},\mathbf{H})$,
and demonstrate how it can be deployed to 
find $\mathbf{H}_\text{opt}$ in Equation \ref{eq:problem_formulation}.

\subsection{An intelligible objective function}
The global objective function $\mathcal{M}(\mathbf{E},\mathbf{H})$ in Equation \ref{eq:problem_formulation}
must favor the selection of dominant equation terms intelligibly, 
where we define intelligible as both 
interpretable (\cite{rudin2019stop}) 
\textit{and} congruent with domain knowledge (\cite{wang2018learning}). 
We construct the objective function 
such that the definition of the 
optimal satisfies two conditions for 
each regime:
\begin{enumerate}
    \item the magnitude difference between the selected dominant terms and the negligible terms must be maximized;
    \item the magnitude difference between the terms within the selected dominant set must be minimized;
\end{enumerate}
and such that the global optimal for possibly many regimes satisfies criteria 1) and 2) for 
the largest weighted percentage of samples. 
If 
the first condition is not satisfied, then all 
equation terms should be retained, i.e., they are all equally dominant.

We begin by defining the local order-of-magnitude score, $\mathcal{M}_n(\mathbf{e}_n,\mathbf{h}_n)$, hereafter the 
local magnitude score,
pertaining to a single observation, 
which measures the magnitude gap between dominant terms $\mathbf{h}_n \cdot \mathbf{e}_n$ and negligible terms $|\mathbf{h}_n-\mathbf{1}| \cdot \mathbf{e}_n$ in a single observation $\mathbf{e}_n$  
(recall that the terms that are selected as dominant are labeled by ${h}_{ni}=1$, and the neglected 
terms are labeled by ${h}_{ni}=0$).
Define $\mathrm{F}=\{1, . . . ,D\}$ as the \textit{index set} (\cite{munkres2000topology}) of the indices of the full set equation terms in vector $\mathbf{e}_n$, such that
\begin{equation}
  \mathbf{e}_n=\bigcup_{i\in \mathrm{F}} e_{ni},
\end{equation}
for observation $n$ such that $1\leq n \leq N$.

We refer to the binary sets that represent the dominant terms as {\it hypotheses}, 
because 
they represent informal 
equation truncations that are not guaranteed to have asymptotic properties. The hypotheses for the entire data set 
$\mathbf{E}$ form an array, $\mathbf{H}$, which has the same dimensions as $\mathbf{E}$, 
namely number of samples $\times$ number of equation terms.
The hypothesis vectors for each observation 
can be expressed as
\begin{equation}
\mathbf{h}_n=\bigcup_{i\in\mathrm{F}} h_{ni},\label{eq:hypothesis_vec}
\end{equation}
where $\mathbf{h}_n$ is an
indicator function (\cite{cormen2009introduction}) that consists entirely 
of ones and zeros, which represent selected dominant terms and negligible 
terms, respectively. 

The indices of elements in $\mathbf{e}_n$ that are selected as dominant
terms by the hypothesis $\mathbf{h}_n$ form the selection index set $\mathrm{S}_n$, where
\begin{equation}
 \mathrm{S}_n \subseteq \mathrm{F} 
.
\end{equation}
Note that the number of selected elements may vary for each observation $n$, and if $\mathrm{S}_n=\mathrm{F}$ then $\mathbf{h}_n=\mathbf{1}$ 
and no equation 
terms are neglected.
It follows that the remainder index set $\mathrm{R}_n$ for the $n^{th}$ observation is 
defined by set subtraction,
\begin{equation}
 \mathrm{R}_n = \mathrm{F}-\mathrm{S}_n 
 ,
\end{equation}
and, therefore, the remainder index set and selected index set 
are non-overlapping,
\begin{equation}
\mathrm{R}_n\cap\mathrm{S}_n = \varnothing 
.
\end{equation}
Thus the cardinality, or size, of the selected 
index set and remainder index set 
are $2\leq  \text{card}(\mathrm{S}_n)\leq D$ and $0\leq \text{card}(\mathrm{R}_n)\leq D-2$, respectively. The lower bound of two 
selected terms is not necessary nor required; we 
impose it because a dominant balance of 
just one term is conceptually ambiguous.
Let the arrays of selected and remainder 
equation terms from 
$\mathbf{e}_n$ be $\mathbf{s}_n$ and $\mathbf{r}_n$, 
respectively. $\mathbf{s}_n$ and $\mathbf{r}_n$ 
are normalized by the smallest element of $\mathbf{e}_n$ and
defined as
\begin{flalign}
   \mathbf{s}_n &= \frac{\bigcup_{i\in\mathrm{S}_n} |{e}_{ni}|}{\min\big(\bigcup_{i\in\mathrm{F}} |{e}_{ni}|\big)},
   \label{eq:sn_def}\\
   \mathbf{r}_n &= \frac{\bigcup_{i\in\mathrm{R}_n} |{e}_{ni}|}{\min\big(\bigcup_{i\in\mathrm{F}} |{e}_{ni}|\big)},
    \label{eq:rn_def}
\end{flalign}
respectively. 
If $\min(\bigcup_{i\in\mathrm{F}} |{e}_{ni}|)=0$, 
then the minimum non-zero absolute valued element of 
$\mathbf{e}_n$ replaces the denominators in Equations 
\ref{eq:sn_def} and \ref{eq:rn_def}.
Let the relative magnitude gap between the normalized subsets, $\Gamma$, 
be defined as a scalar 
for each $n^{th}$ observation: 
\begin{equation}
\Gamma_n=\begin{cases} 
\frac{ \log_{10}(\min(\mathbf{s}_n) -  \max(\mathbf{r}_n)) }
{  \log_{10}(\min(\mathbf{s}_n) + \max(\mathbf{r}_n)) } 
\quad \text{if} \quad \min(\mathbf{s}_n)>\max(\mathbf{r}_n)\\
0 \hspace{35.5mm} \text{if} \quad \min(\mathbf{s}_n)\leq\max(\mathbf{r}_n)
\end{cases} .
\end{equation}
The magnitude gap $\Gamma$ is normalized such that $\Gamma\in[-\infty,1]$, and the floor condition 
\textit{if $\Gamma<0$ then set $\Gamma=0$} is implemented to correct for spurious large amplitude 
negative values of 
$\Gamma$ that arise as $\min(\mathbf{s}_n)\rightarrow\max(\mathbf{r}_n)$ 
from above.
$\Gamma\rightarrow1$ as the number of orders of magnitude between the element with the minimum absolute value of the selected subset and element with the maximum absolute value of the remainder subset 
approaches infinity, and if $\Gamma=1$ then the hypothesis
that the terms in the selected subset dominate the terms in the remainder subset 
is exact (for any numerical implementation the optimal is limited by machine precision, so $\Gamma\approx1$).
If the magnitude difference 
between the two subsets vanishes, then $\Gamma\rightarrow0$, and 
if a remainder subset term exceeds the absolute magnitude of the 
selected subset, then the hypothesis does not represent a dominant 
balance, and $\Gamma=0$ is prescribed.

Since the goal is to choose the selected subset, $\mathbf{s}_n$, such that it corresponds to the dominant terms, the feature magnitudes of the selected subset 
should be approximately the same. 
Otherwise, the smallest magnitude term(s) in the selected subset should be removed from 
that subset and added to the remainder subset. 
To penalize large absolute magnitude differences within the selected 
subset, we introduce
the scalar penalty for the $n^{th}$ observation, 
\begin{equation}
\Omega_n=\log_{10}(\max(\mathbf{s}_n)) -  \log_{10}(\min(\mathbf{s}_n))\in[0,\infty).
\end{equation}
A base 10 logarithm is chosen for the penalty because it corresponds most directly 
to the notion of orders of magnitude.
If $\Omega_n\rightarrow0$, the absolute magnitudes of the selected subset terms 
approach uniformity. 

We define the local magnitude score for the $n^{th}$ observation $\mathcal{M}_n(\mathbf{e}_n,\mathbf{h}_n)$ 
as 
\begin{equation}
\mathcal{M}_n(\mathbf{e}_n,\mathbf{h}_n)=\frac{\Gamma_n}{1+\Omega_n}\in[0,1].\label{eq:M_n}
\end{equation}
By construction, the local magnitude score is a) invariant to the magnitude of the feature vector $\mathbf{x}_n$ 
 and b) invariant to the sign of the elements of the feature vector,
 \begin{equation}
      \mathcal{M}_n(\mathbf{e}_n,\mathbf{h}_n)
      =\mathcal{M}_n(\pm c\mathbf{e}_n,\mathbf{h}_n),\label{eq:invariant_mult}
 \end{equation}
 where $c$ is a positive scalar constant. Therefore, 
 the score is invariant to the choice of dimensional 
 or non-dimensional equations, and, equivalently, it can be applied to Buckingham $\Pi$ theorem
 to identify dominant $\Pi$ groups. 
 Readers unfamiliar with the 
 Buckingham $\Pi$ theorem and how 
 it pertains to the non-dimensionalization of partial differential equations are referred to \cite{zohuri2017dimensional}.

Finally, we propose the global magnitude score $\mathcal{M}(\mathbf{X},\mathbf{H})$
as an objective function that satisfies Equation \ref{eq:problem_formulation},
which yields a single score for the fit of the hypotheses for 
all $N$ observations, $\mathbf{H}$, to the 
entire data set $\mathbf{E}$,
\begin{equation}
\mathcal{M}(\mathbf{E},\mathbf{H})= \frac{\sum_{n=1}^N {w}_n\cdot  \mathcal{M}_n(\mathbf{e}_n,\mathbf{h}_n)}{\sum_{n=1}^N{w}_n},
\label{eq:global_score}
\end{equation}
where the array of weights $\mathbf{w}=[w_1,...,w_N]$ are the discrete differentials of 
the observed domain, e.g. space and/or time differentials. For example, if the $N$ observations of 
data set $\mathbf{E}$ are of equation terms distributed across a two-dimensional space, then the global magnitude 
score in Equation \ref{eq:global_score}
is the area-weighted average of all of the scores for each observation. 
The full set score for the entire data set can be expressed as $\mathcal{M}(\mathbf{E},\mathbf{1})$, where 
$\mathbf{1}=\{\mathbf{1}_1,...,\mathbf{1}_N\}$. 
Equation \ref{eq:global_score} is an 
evaluation of the fit of 
potentially many different hypotheses to 
potentially many different regimes within 
data set $\mathbf{E}$.

\subsubsection{Combinatorial hypothesis selection} 
We propose a simple hypothesis
selection algorithm 
that we will refer to as the 
combinatorial hypothesis selection 
(CHS) algorithm. Since the number of all possible 
hypotheses for an equation is a permutation of 
two types (0 or 1) with repetition allowed, the 
number of possible hypotheses is $\mathcal{O}(2^D)$. If the number of 
equation terms, $D$, is not large, then hypotheses can 
be feasibly generated by calculating the 
magnitude score (Equation \ref{eq:M_n}) for all possible hypotheses  
and then selecting the hypothesis 
that is awarded the highest score. Equation \ref{eq:M_n} 
can be applied to a single data sample or to an average of samples.
The exponential time complexity limits the feasibility of computing 
CHS to 
equations with relatively few terms. 

\subsection{Unsupervised learning framework}
In unsupervised learning,
the goal is to partition the data into labeled groups, or clusters,
that reveal underlying patterns of sparsity in the data. 
Clustering algorithms are a class of unsupervised machine learning algorithms that yield a finite set of categories to describe a given data set according to similarities or relationships among its objects 
(\cite{macqueen1967some}, \cite{hartigan1975clustering}).
Clustering performance scores are objective measures that define 
the optimal partitioning of the data by 
quantitatively evaluating the consistency of the statistical properties of the 
clusters with hypothesized statistical properties of the data.
For example, if one wishes to 
find convex Euclidean (hyper-spherical or hyper-elliptical) 
clusters, then the 
silhouette score (\cite{rousseeuw1987silhouettes}) 
is a useful measure of the fit of
the clustering algorithm labels to
convex groups of features that may 
present in the data. Clustering performance scores 
are particularly useful because
clustering results are sensitive to the choice of 
algorithm parameters (\cite{scikit-learn}) 
and there is no definition of a cluster that is universal to all clustering algorithms (\cite{estivill2002so}).

\begin{figure}[h!]
 \centering
 \includegraphics[width=0.85\textwidth]{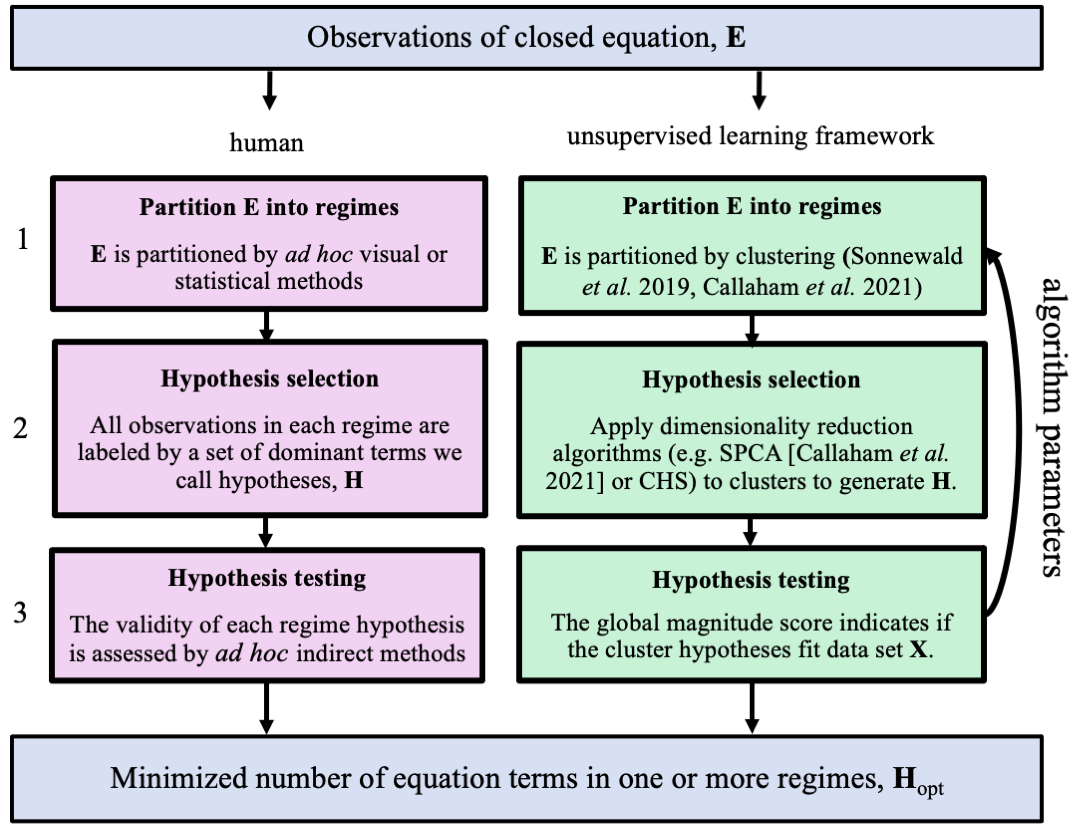}
 \caption{Diagram of the dynamical regime problem, with 
 partitioning and empirical scaling analysis performed 
 by a human (left column), and algorithms capable of 
 performing said tasks (right column). The  
 loop over algorithm parameters illustrates the 
 procedure for obtaining $\mathbf{H}_\text{opt}$ 
 in Equation \ref{eq:problem_formulation}.}
 \label{fig:algorithm}
 \end{figure}

The objective score $\mathcal{M}(\mathbf{E},\mathbf{H})$ in Equation \ref{eq:global_score}
is a quantitative measure of 
the sparsity of equation terms and
a criterion for a framework that includes both
clustering and hypothesis selection algorithms.
Given equation data set $\mathbf{E}$,  
clustering algorithms can be coupled with hypothesis selection 
methods to solve the dynamical regime problem by labeling 
regions of $\mathbf{E}$ with dominant balance hypotheses $\mathbf{H}$ (\cite{callaham2021learning}). 
We propose a 
framework for unsupervised learning (\cite{kohavi1997wrappers}, \cite{dy2004feature}) 
that utilizes the objective criterion $\mathcal{M}(\mathbf{E},\mathbf{H})$ 
to loop over algorithm parameters 
to find optimal dominant balance hypotheses, $\mathbf{H}_\text{opt}$, as defined by Equation \ref{eq:problem_formulation}.
The framework, akin to the scientific method, is depicted in 
Figure \ref{fig:algorithm}, in which
the dynamical regime problem is broken 
into partitioning, 
hypothesis selection, and hypothesis testing tasks.
The left column describes the historical heuristic method for 
solving the dynamical regime problem and the right column depicts our algorithmic 
framework.

\subsubsection{Partitioning dynamical regimes}
The first task shown in Figure \ref{fig:algorithm}, row 1, is to classify different dynamical regimes within the feature data (equation terms) with a clustering algorithm. 
For humans, this task is a simple pattern recognition task of 
merely noticing markedly different dynamics 
in one spatio-temporal region versus another: the act of observing and 
discerning distinct dynamical 
regimes. 
\cite{sonnewald2019unsupervised} suggested that if 
the governing equations of the observed dynamics are known, 
then the heuristic act of 
noticing different dynamical regimes can 
be formulated as
a partitioning task 
that can be performed by a clustering algorithm. 
Dynamical regimes are conventionally defined by dominant \textit{average} 
balances, where the averaging is applied to the sample space (e.g. 
space and/or time).
Parametric clustering algorithms like Expectation-Maximization (EM, \cite{bishop2006pattern}) 
clustering algorithms, which assume that 
the data are drawn from known parametric families of distributions 
(e.g. Gaussian), are consistent with this 
precedent, but the cluster means and variances suffer from a 
sensitivity to outliers 
(\cite{campello2015hierarchical}). Outliers can distort the 
parametric model density functions by 
\textit{masking} or \textit{swamping} 
them
to produce erroneous model means and 
to erroneously enlarged or reduced model variances (\cite{pearson1936efficiency}). 
While nonparametric clustering 
algorithms detect and remove outliers (\cite{hartigan1975clustering}, \cite{campello2015hierarchical}), the optimal choice of 
clustering algorithm for the dynamical regime problem
is not obvious and will depend on the properties of 
both the dynamical information and the noise in each data set. 
The equation data should be standardized in accordance with 
the choice of clustering algorithm.
Crucially, the objective criterion 
$\mathcal{M}(\mathbf{E},\mathbf{H})$ 
defines the optimal for a given clustering algorithm, 
\textit{and} it permits objective comparisons between 
the optimal results of different clustering algorithms.

\subsubsection{Hypothesis selection}
The second task shown in Figure \ref{fig:algorithm} 
is to generate hypotheses $\mathbf{H}$ for all samples. 
Humans typically
perform this task by carefully estimating characteristic 
scales from observations, subsequently estimating the magnitude of non-dimensional parameters, and finally choosing
which equation terms are negligible 
in each regime (\cite{zohuri2017dimensional}).
The goal of hypothesis selection 
is to assign the same hypotheses 
to all samples within a regime.
This is achieved by applying statistical
dimensionality reduction techniques to
assign
active or inactive (negligible) labels 
to the equation terms uniformly for all samples 
within each cluster.
Therefore, 
hypothesis selection 
algorithms are merely algorithms selected from  
the subclass of
dimensionality reduction techniques that pertain to convex data (\cite{van2009dimensionality}).
This was first demonstrated in the context of equation-spaces by \cite{callaham2021learning}, who proposed 
sparse principal component analysis (SPCA, \cite{zou2006sparse}) as a tenable hypothesis 
selection algorithm because it labels features 
with small variances as negligible by performing a
least absolute shrinkage and selection operator (LASSO, \cite{tibshirani1996regression}) regression on the principal axes 
from principal component analysis. SPCA is therefore 
notionally consistent with EM clustering algorithms 
(e.g. $K-$means, Gaussian Mixture Models) in the sense that 
both clustering algorithm and hypothesis selection algorithm 
are parametric and 
specifically assume that the sampled cluster feature variances are representative of 
a unimodal Gaussian or near-Gaussian cluster feature distributions.

\subsubsection{Hypotheses testing}
 The final task shown in Figure \ref{fig:algorithm} 
 is to evaluate the fit of the hypotheses to each regime.
 In the past, this task has been performed both directly 
 and indirectly.
 If the observations included not only enough data to estimate the 
 characteristic scales but also enough data to estimate the 
 amplitudes of the equation terms, then direct evaluations of 
 which equation terms are negligible could be performed. Otherwise, 
 hypotheses were indirectly validated through the validation of 
 predictive models that neglected the same equation terms 
 (e.g. aerodynamic drag forces predicted by the ``law of the wall'').
 However, in the case of direct validation, an arbitrary 
 threshold that separates dominant from negligible terms must be selected. 
 The threshold must be applied locally and not globally, meaning it 
 must be applied to each regime individually, 
 because the absolute 
 equation term magnitudes can vary from regime to regime, as
 exemplified by d'Alembert's paradox (see \textit{Introduction}, also \cite{currie2016fundamental}). 
 Therefore, by defining 
 the optimal hypotheses for the dynamical regime problem, Equation \ref{eq:problem_formulation} formalizes direct validation of 
 hypotheses. 
 The framework loop over clustering and hypothesis selection
 algorithm parameters depicted in Figure \ref{fig:algorithm}
 is thus conceptually akin to a human applying different 
 characteristic scales and/or methods of estimating characteristic scales
 while searching for the optimal hypotheses for a given data set.

\section{Examples}
In this section we demonstrate the use of the local magnitude score and the unsupervised 
learning framework to solve the problem of identification of dynamical regimes, using examples drawn from a variety of disciplines. Complete 
discussions of the dynamics of each example are beyond the scope of this Article 
and interested readers are referred to the corresponding citations.
\subsection{Local magnitude score example} 
 \begin{figure}[b!]
 \includegraphics[width=1.\textwidth]{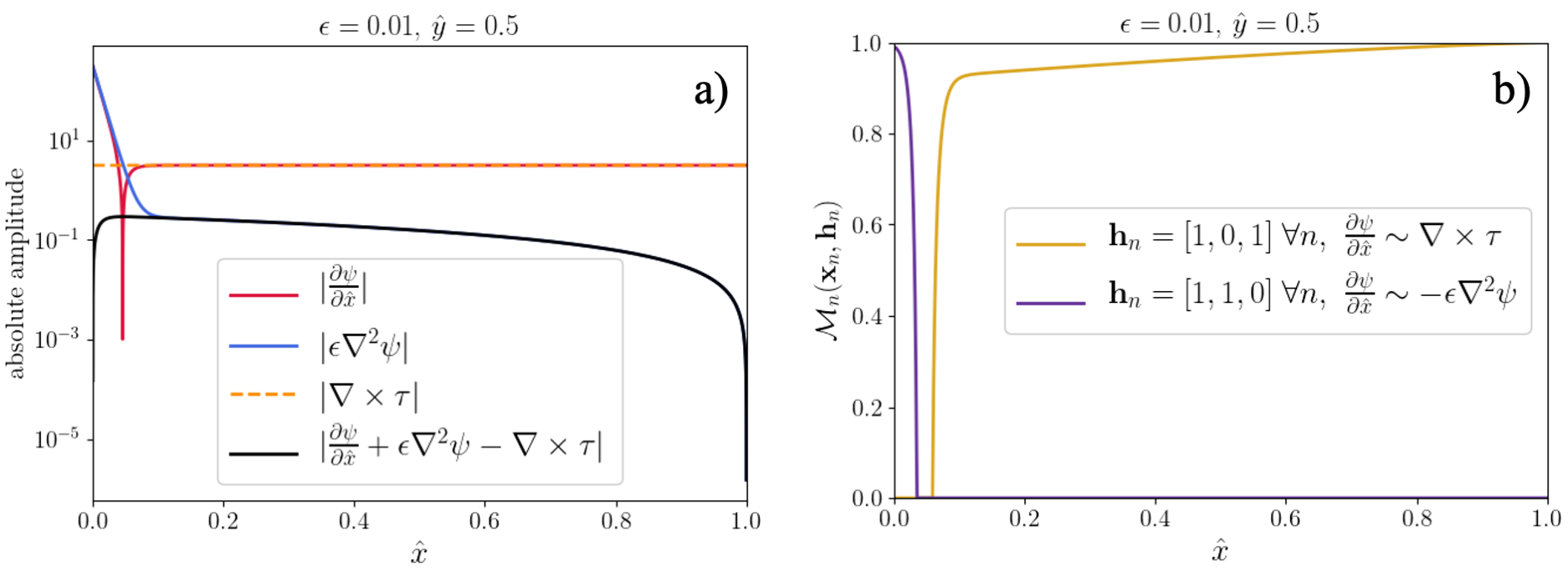}
 \caption{The local magnitude score applied to 
 Munk's asymptotic matching problem for wind-driven ocean circulation, using the low-order diffusion model of \cite{vallis2017atmospheric}. Panel $a)$ shows the magnitude of 
 each equation term as a function of $\hat{x}$ if $\hat{y}=0.5$, 
 and panel $b)$ shows the local magnitude for two hypotheses, the purple line corresponding to 
 a western boundary current balance and the 
 yellow line corresponding to Sverdrup balance.}
 \label{fig:munk}
 \end{figure}
Consider the asymptotic 
matching problem presented by \cite{munk1950wind} 
in his analysis of the dynamics 
of wind-driven oceanic heat transport (Figure \ref{fig:munk_1950}).
The two-dimensional 
steady-state circulation is described by 
its non-dimensional streamfunction 
$\psi(\hat{x},\hat{y})$, 
forced by the non-dimensional wind stress 
$\boldsymbol{\tau}=-\cos(\pi \hat{y})\mathbf{i}$, 
and governed by the equation
\begin{equation}
    \overbrace{\frac{\partial\psi}{\partial\hat{x}}}^{\substack{\text{advection of}\\\text{ planetary vorticity}}}
    \hspace{-5mm}+\hspace{3mm}\overbrace{\epsilon\nabla^2\psi}^{\substack{\text{diffusive}\\\text{torque}}}\hspace{2mm}=\hspace{2mm}
    \overbrace{\nabla\times\boldsymbol{\tau}}^{\substack{\text{wind}\\\text{stress curl}}},
\label{eq:vallis}
\end{equation}
where $0<\epsilon\ll1$, $\nabla^2=\partial^2/\partial\hat{x}^2+\partial^2/\partial\hat{y}^2$, $\hat{x}$  
and $\hat{y}$ are 
the longitudinal and latitudinal coordinates, 
respectively,
the ocean basin geometry 
is idealized as a square, $0\leq \hat{x} \leq 1$, $0\leq \hat{y} \leq 1$, and we employ the lower-order 
diffusive torque model 
of \cite{vallis2017atmospheric} for illustrative purposes.
The method of matched 
asymptotic expansions can be applied to obtain 
the solution (\cite{vallis2017atmospheric}) 
\begin{equation}
    \psi(\hat{x},\hat{y})\approx(1-x-\mathrm{e}^{-\hat{x}/\epsilon})\pi\sin(\pi \hat{y}),\label{eq:munk_asymptotic}
\end{equation}
which approximately satisfies 
impermeable flow boundary conditions ($\psi=0$) 
on both the 
east and west sides of the ocean basin. The solution 
to this problem, and therefore each possible dynamical regime or 
truncation of Equation \ref{eq:vallis}, 
are referred to as asymptotic because the conservation properties converge consistently with $\mathcal{O}(\epsilon)$. However, Equation \ref{eq:vallis} is an idealized and 
truncated form of the barotropic vorticity 
equation, and therefore Equation \ref{eq:munk_asymptotic} is not 
an asymptotic solution of the full vorticity equation.

Figure \ref{fig:munk}a) shows the 
magnitude of each equation term in 
the asymptotic matching problem
for $\epsilon=0.01$ at $\hat{y}=0.5$, such 
that each $n^{th}$ observation $\mathbf{e}_n=[\frac{\partial\psi}{\partial\hat{x}},\epsilon\nabla^2\psi,\nabla\times\boldsymbol{\tau}]_n$ 
corresponds to a discrete location in the $x$ direction. 
Figure \ref{fig:munk}b) shows the 
variation of
local magnitude score over $\hat{x}$ at $\hat{y}=0.5$ 
for two dominant balances:
a dominant balance between planetary advection and 
diffusion (a.k.a. western boundary current, $\mathbf{h}_n=[1,0,1]$, purple) and a dominant balance between 
the planetary advection and the wind stress curl 
(a.k.a. Sverdrup balance, $\mathbf{h}_n=[1,1,0]$, yellow). 

Figure \ref{fig:munk}b) indicates that, at $\hat{y}=0.5$,
the western boundary current and Sverdrup dynamical 
regimes are unambiguously dominant over distinct ranges of 
$\hat{x}$ separated by a gap at $\hat{x}\approx0.04$. 
However, the boundaries 
between different dynamical regimes are often not as obvious as in Figure \ref{fig:munk}b) and an objective method for partitioning 
the regimes is required.
In the next section, we demonstrate how 
observations can be 
partitioned into distinct regimes 
by using the 
unsupervised learning methods of \cite{sonnewald2019unsupervised} and 
\cite{callaham2021learning}, which 
can then be evaluated using the objective 
function defined by Equation \ref{eq:problem_formulation} to find the optimal partitioning 
and labeling of each dynamical regime.

\subsection{Unsupervised learning framework examples}
\subsubsection{Synthetic data}
\label{sec:synth_data}
A simple framework 
test case is a two-dimensional array of data with an even number of features 
where half of the features 
are many orders of magnitude larger in one half of the domain and vice versa. Multiplicative sinusoidal 
noise is added to give the two regions variance that is proportional to 10\% of the signal 
amplitude in each region. The two regions can be considered dynamical regimes: in 
each, half of the terms dominate the other equation terms by two orders of magnitude. 
The regions 
of dominant terms are prescribed by a perturbed Heaviside step function $\mathcal{H}$, such that:
\begin{flalign}
    \mathbf{e}_i(\hat{x},\hat{y})&=(-1)^i {\eta}(\hat{y}) ( \lambda\mathcal{H}(\phi) + \beta),\label{eq:synth1}\\
    \eta(\hat{y})&=\eta_0\sin(\omega \hat{y}),\\
    \phi&=\begin{cases} \hat{y}-0.5 \quad \forall  i< D/2\\
    0.5-\hat{y}\quad \forall  i\geq D/2
    \end{cases}\label{eq:synth3},
\end{flalign}
where $\hat{x}$ and $\hat{y}$ are spatial coordinates.
The equation closes exactly for all $n$ samples, $\sum_{i=1}^D e_{ni} = 0$,
and the prescribed coefficients are shown in Table \ref{table:1}. 

\begin{table}[h!]
\centering
\begin{tabular}{ |c|c|c|c|c| } 
\hline
 $\lambda$ & $\beta$ & $\eta_0$ & $\omega$ \\
\hline
 $10^{1}$ & $10^{-1}$ & $10^{-1}$ & $10\pi$ \\
\hline
\end{tabular}
\caption{Prescribed coefficients for the synthetic data set}
\label{table:1}
\end{table}

\begin{figure}[h!]
 \includegraphics[width=1.\textwidth]{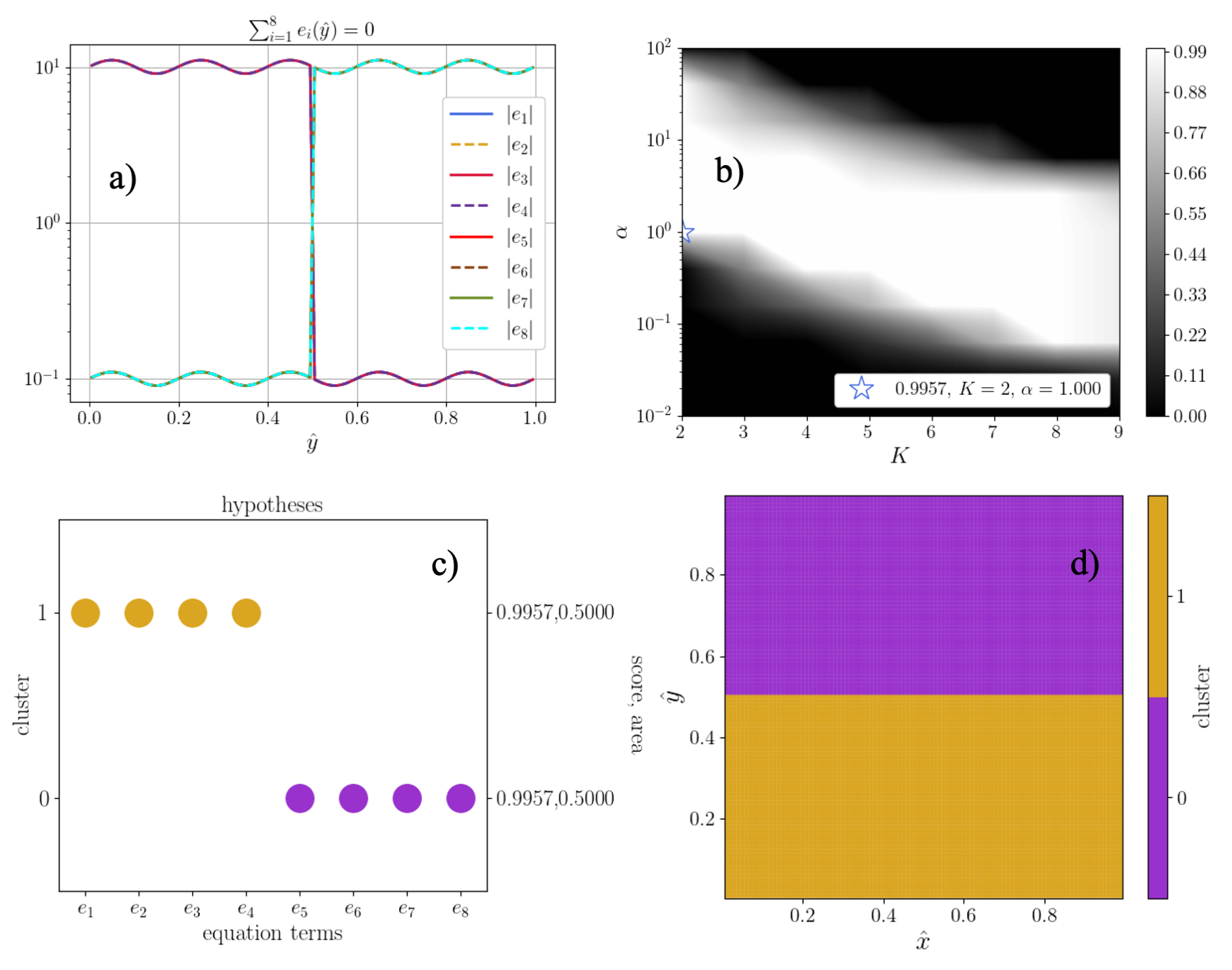}
 \caption{The synthetic data for all $\hat{y}$ 
 at constant $\hat{x}$ is 
 shown in panel $a)$. Panel $b)$ shows the variation global magnitude score as 
 the number of prescribed clusters for $K-$means and the LASSO 
 regression coefficient $\alpha$ for SPCA are varied. Panels $c)$ and $d)$ show 
 the dominant balances and their spatial distribution  for the 
 optimal results that occur in the white band in $b)$ which 
 correspond to a global magnitude score of 0.996.}
 \label{fig:synth}
 \end{figure}

Figure \ref{fig:synth}a) 
shows the synthetic data $e_{ni}$ consisting of $D=8$ equation terms and 
featuring
two regimes in which the dominant terms
have
amplitudes of $\mathcal{O}(10)$.
Each regime is composed of four equation terms, 
and the inactive terms have amplitudes of   $\mathcal{O}(10^{-1})$, and the regimes are
separated by a discontinuity at $\hat{y}=0.5$ as 
prescribed by Equation \ref{eq:synth1}.

Figures \ref{fig:synth}b), \ref{fig:synth}c), and \ref{fig:synth}d) show the 
unsupervised learning framework results for 
the synthetic data using
$K-$means for clustering and SPCA for hypothesis selection.
Figure \ref{fig:synth}b) shows 
the variation of the global magnitude score $\mathcal{M}(\mathbf{E},\mathbf{H})$ as
$\alpha$, the LASSO regression coefficient 
for SPCA, and $K$, the prescribed number 
of clusters for $K-$means clustering, are varied. The 
optimal result is marked with the blue star.
Figure \ref{fig:synth}c) shows the dominant equation terms 
of the optimal result
and Figure \ref{fig:synth}d) shows the spatial distribution of the
two clusters of the optimal result. 
The clusters and their distributions for all of 
the $\alpha$ and $K$ values within the white 
band where the global magnitude score is 
$\mathcal{M}(\mathbf{E},\mathbf{H})=0.9957$ in 
Figure \ref{fig:synth}b) are identical to those 
shown in Figures \ref{fig:synth}c) and 
\ref{fig:synth}d). 

The same optimal results as shown in Figures \ref{fig:synth}b), \ref{fig:synth}c), and \ref{fig:synth}d) were obtained 
by using $K-$means clustering and CHS, 
by using Hierarchical Density-Based Scan (HDBSCAN) 
and SPCA, as well as by using HDBSCAN and the 
CHS algorithm. 
The robustness of the results shown in Figures \ref{fig:synth}b), \ref{fig:synth}c), and \ref{fig:synth}d) arises because the magnitude separation 
and spatial distributions of the dominant balances 
that are implicit in the synthetic data (see Figure \ref{fig:synth}a)) are  
pronounced in both amplitude and the sharpness of the 
boundary. 
The results may be sensitive to the choice of clustering and hypothesis selection algorithms for data sets that contain mixed order 
balances and smoothly varying dominant balances.
However, the global score (Equation \ref{eq:global_score}) allows for objective 
comparisons of algorithm choices such that the optimal 
algorithms can be selected for any given data set.
To summarize, Figure \ref{fig:synth} shows that the framework (Figure \ref{fig:algorithm}) and the proposed objective function (Equation \ref{eq:global_score}) yield robust solutions to Equation \ref{eq:problem_formulation}, independent of the 
chosen algorithms, for data sets in which the boundaries of 
dynamical regimes are sharp and the dominant balances within 
the regimes dominate by at least two orders of magnitude.

\subsubsection{Global ocean barotropic vorticity}
\begin{figure}[h!]
\centering
\includegraphics[width=1.0\textwidth]{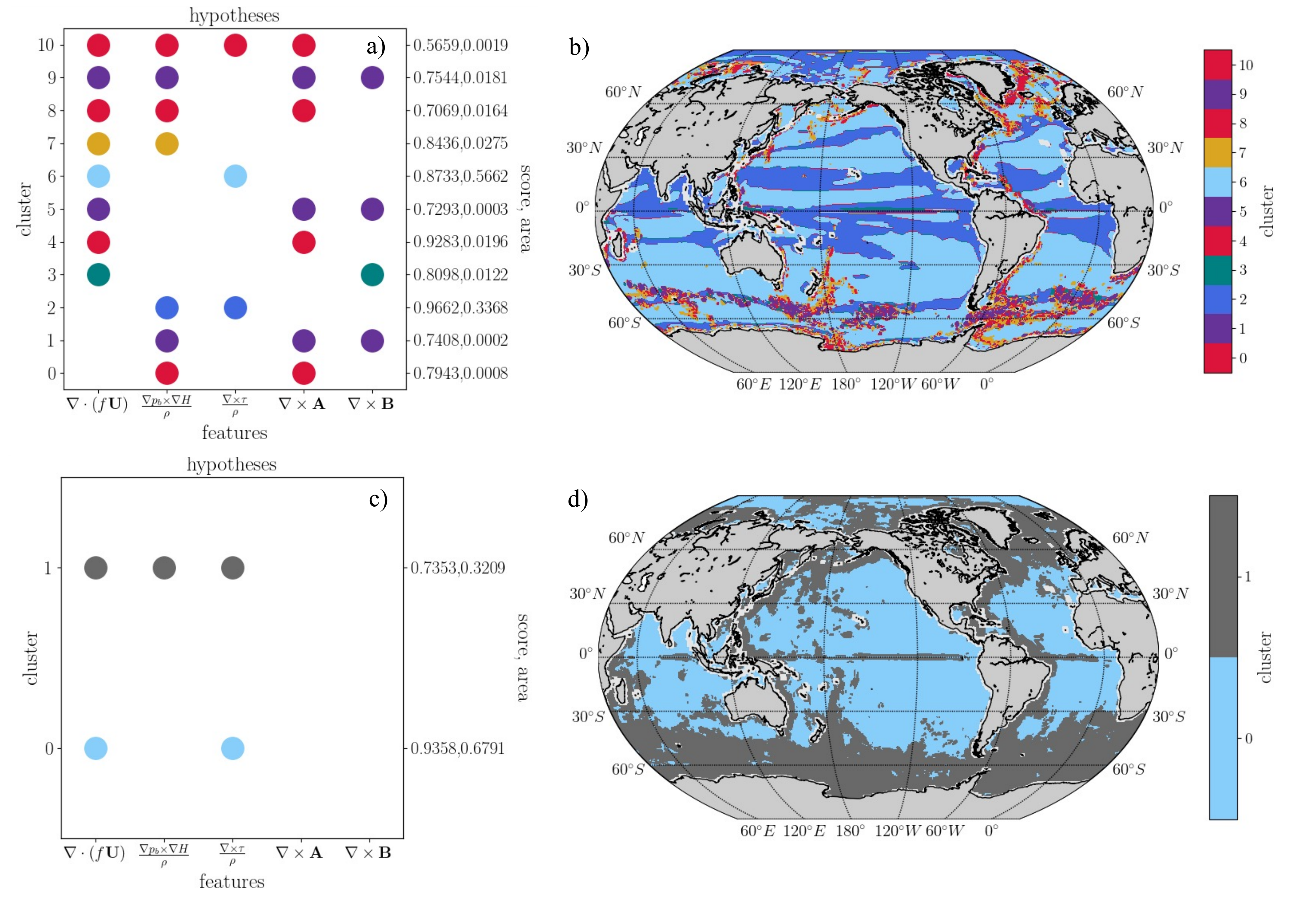}
\caption{Panels $a)$ and $b)$ show the optimal dominant balances for each 
cluster and their spatial distributions, respectively, for $K-$means 
clustering and CHS algorithms. The optimal score 
$\mathcal{M}(\mathbf{E},\mathbf{H})=0.90$ was found at $K=49$. 
Panels $c)$ and $d)$ show the optimal dominant balances for each 
cluster and their spatial distributions, respectively, for HDBSCAN 
clustering and CHS algorithms. The optimal score 
$\mathcal{M}(\mathbf{E},\mathbf{H})=0.87$ was found for 
50 minimum samples and 200 minimum cluster size. 
The $K-$means results score higher than the HDBSCAN and 
they agree with the analysis of \cite{sonnewald2019unsupervised}.
}
\label{fig:ecco}
\end{figure}
\cite{sonnewald2019unsupervised} calculated the terms in the barotropic vorticity equation for the global ocean, and subsequently 
used the $K-$means algorithm to partition them into dynamical regimes that were  qualitatively 
interpreted as dominant balances. Since the $K-$means and CHS algorithms and 
SPCA all implicitly assume that convex, unimodal clusters are 
present in the data, and the number of equation terms is small, 
we chose the $K-$means clustering with the CHS algorithm to 
test the framework shown in Figure \ref{fig:algorithm} on 
the barotropic vorticity problem of \cite{sonnewald2019unsupervised}.
They computed a 
20-year mean of the Estimating the Circulation and Climate of the Ocean (ECCO)
ocean state estimate version 4 release 2 (see \cite{forget2015ecco}, \cite{wunsch2013dynamically}, \cite{ecco2017twenty1}, and \cite{ecco2017twenty2}) at 1$^\circ$ resolution 
to calculate terms of the vertically integrated barotropic vorticity equation, which 
was grouped and rearranged to be expressed as
\begin{equation}
     \overbrace{\nabla\cdot(f\mathbf{U})}^{\substack{\text{advection of}\\\text{ planetary vorticity}}}\hspace{-5mm}=
\hspace{2mm}\overbrace{\frac{\nabla p_b \times \nabla H}{\rho}}^{\substack{\text{bottom}\\\text{pressure torque}}}\hspace{3mm}
+\hspace{-2mm}\overbrace{\frac{\nabla\times\boldsymbol{\tau}}{\rho}}^{\substack{\text{wind \& bottom}\\\text{stress curl}}}\hspace{-2mm}
+\hspace{1mm}\overbrace{\nabla\times\mathbf{A}}^{\substack{\text{nonlinear}\\\text{torque}}}\hspace{1mm}
+\hspace{1mm}\overbrace{\nabla\times\mathbf{B}}^{\substack{\text{
diffusive}\\\text{torque}}}.
\label{eq:BVeqn}
\end{equation}
In the above, $f$ is the Coriolis parameter, $\mathbf{U}$ is 
the vertically integrated horizontal velocity, $p_b$ 
is the bottom pressure, $H$ is the depth, $\rho$ is a 
reference density, $\boldsymbol{\tau}$ represents surface 
stress, $\nabla$ is applied only to the horizontal coordinates, 
$\mathbf{A}$ contains nonlinear horizontal momentum fluxes, and $\mathbf{B}$ contains linear horizontal diffusive fluxes.

Figures \ref{fig:ecco}$a)$ and \ref{fig:ecco}$b)$ show the optimal dominant balances for each
cluster and their spatial distributions, respectively, for $K-$means 
clustering and CHS algorithms. The optimal score 
$\mathcal{M}(\mathbf{E},\mathbf{H})=0.90$ was found at $K=49$. 
Figures \ref{fig:ecco}$c)$ and \ref{fig:ecco}$d)$ show the optimal dominant balances for each 
cluster and their spatial distributions, respectively, for HDBSCAN 
clustering and CHS algorithms. The optimal score 
$\mathcal{M}(\mathbf{E},\mathbf{H})=0.87$ was found for 
50 minimum samples and 200 minimum cluster size. 
Since the $K-$means results score higher than the HDBSCAN results, $K-$means is the better choice 
of clustering algorithm for this problem. 
Figures \ref{fig:ecco}$b)$ and \ref{fig:ecco}$d)$ also show that results that differ only slightly in score 
can differ in the distribution and labeling of dynamical regimes significantly. 
This underscores the obligation of the user to choose clustering algorithms that 
are constructed upon assumptions that are consistent with the notion of a dynamical regime 
(e.g. dominant mean balances) and that the user must judiciously search parameter 
space for optimal scores. Crucially, our method allows the user to find the best results 
(e.g. Figures \ref{fig:ecco}$a)$ and \ref{fig:ecco}$b)$ relative to Figures \ref{fig:ecco}$c)$ and \ref{fig:ecco}$d)$) without relying \textit{a priori} dynamical knowledge, and the method is applicable to all clustering algorithms and dimensionality 
reduction algorithms and therefore allows for comparisons of the selected algorithms.

The dominant balances shown in Figures \ref{fig:ecco}$a)$ and \ref{fig:ecco}$b)$ are qualitatively consistent with 
and quantitatively similar to the analysis of \cite{sonnewald2019unsupervised}. In particular, 
the distribution of the Sverdrup balance (cluster 6, 
light blue, a balance of planetary advection and wind stress curl) are consistent with the largest magnitude cluster-average terms in \cite{sonnewald2019unsupervised} in the same areas, as well as with domain knowledge. The pressure torque $-$ 
wind stress curl balance (cluster 2, blue) and 
the nonlinear clusters (purple and red clusters) also 
agree with \cite{sonnewald2019unsupervised}.

\subsubsection{Reaction-diffusion dynamics in 
tumor-induced angiogenesis}
A common approach to evaluate the dynamical significance of individual 
equation terms is to calculate numerical solutions with different permutations of 
terms eliminated.
This technique has been used to study tumor angiogenesis (\cite{anderson1998continuous},  \cite{anderson2000mathematical}),
the patterns of growth of 
cells that facilitate the creation of a pathway for blood to reach a tumor thus enabling tumor growth. We demonstrate that our framework provides a direct evaluation of 
which terms are dominant without iterating over different eliminated terms. 

The continuous tumor-induced angiogenesis model of \cite{anderson1998continuous} 
is composed of conservation laws of three continuous variables: 
the endothelial-cell density per 
unit area $n$ (cells that rearrange and migrate from preexisting vasculature to 
ultimately form new capillaries), the tumor angiogenic factor concentration $c$ (chemicals secreted by 
the tumor that promote angiogenesis), and the 
fibronectin concentration $f$ (macromolecules that are secreted 
by $n$ and stimulate the directional migration of $n$). 
Endothelial cell migration up the fibronectin concentration gradient 
is termed haptotaxis (\cite{carter1965principles}, \cite{carter1967haptotaxis}), 
while endothelial cell migration 
up the gradient of tumor angiogenic factor concentration is termed 
chemotaxis (\cite{sholley1984mechanisms}). 

\begin{figure}
\centering
\includegraphics[width=1.0\textwidth]{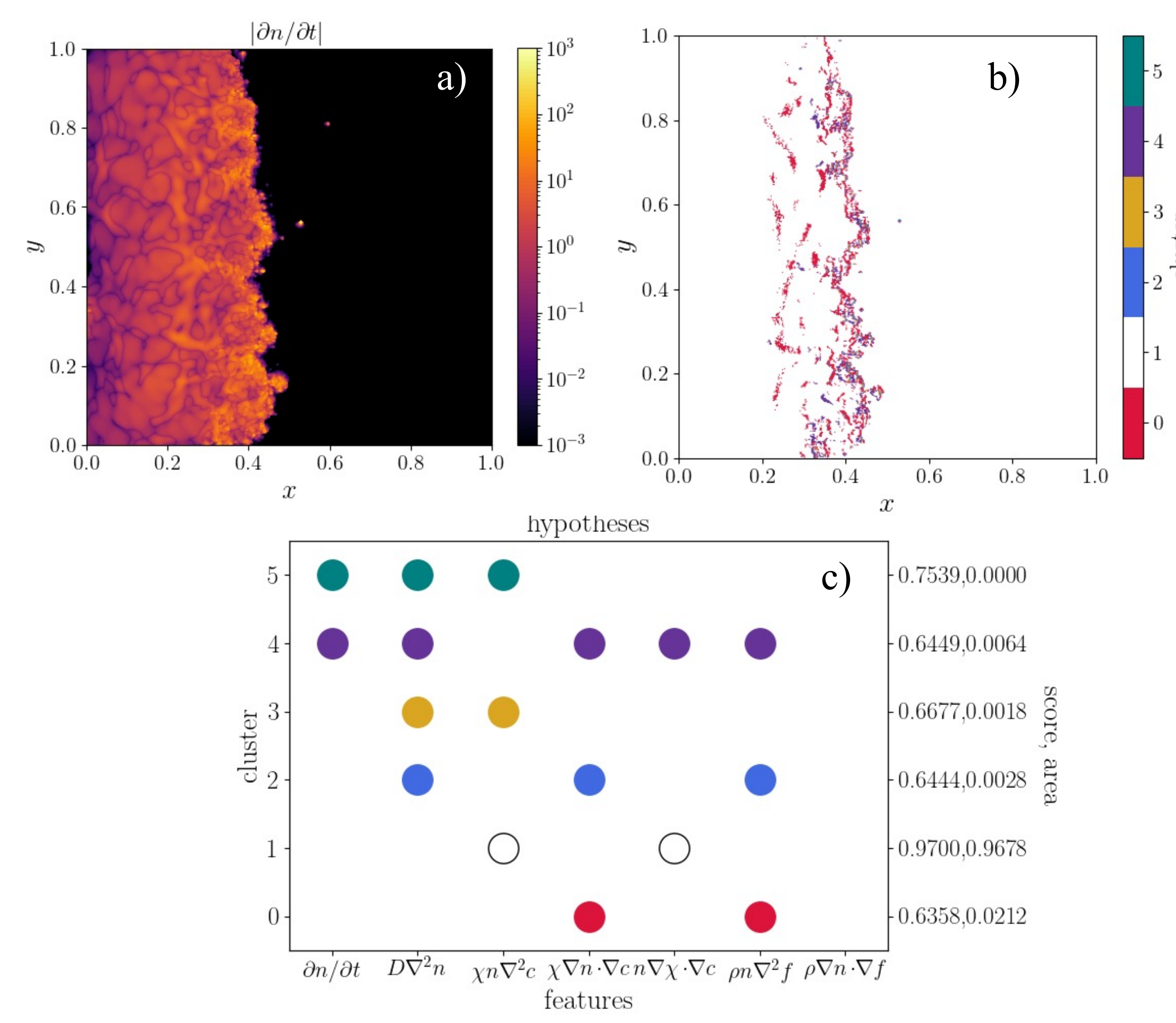}
\caption{Dominant balance discovery for the nonlinear 
reaction-diffusion
tumor angiogenesis problem of \cite{anderson1998continuous}, 
shown at non-dimensional time $t=0.91$. In panel $a)$, the 
tumor is located at $x,y=1,0.5$, and the endothelial cell 
growth is propagating in the positive $x$ direction 
towards the tumor. Panels $b)$ and $c)$ show the 
regime distributions and dominant balances, respectively, 
for the optimal results for $K-$means clustering and 
CHS hypothesis selection.
}
\label{fig:tumor}
\end{figure}

The non-dimensional, tumor-induced angiogenesis governing equations of 
\cite{anderson1998continuous} are
\begin{flalign}
  \frac{\partial{n}}{\partial{t}}&=\hspace{-5.5mm}\overbrace{D_a\nabla^2n}^{\text{random motility}}\hspace{-3mm}
  -\hspace{1mm}\overbrace{\nabla\cdot (\chi n\nabla c)}^{\text{chemotaxis}}\hspace{1mm}
  -\hspace{1mm}\overbrace{ \nabla\cdot (\rho_a n\nabla f)}^{\text{haptotaxis}},
  \label{eq:EC_density}\\
  &=D_a\nabla^2n\hspace{1mm}-\chi n \nabla^2c - \chi \nabla n \cdot \nabla c - n \nabla\chi \cdot \nabla c
  -\rho_a n \nabla^2f - \rho_a \nabla n \cdot \nabla f,
  \label{eq:expanded_EC_density} \\
  \frac{\partial{f}}{\partial{t}}&=\beta_f n - \gamma n f,
  \label{eq:fibronectin_concentration}\\
  \frac{\partial{c}}{\partial{t}}&=-\eta_c cn,
  \label{eq:AF_concentration}
\end{flalign}
where
\begin{equation}
    \chi(c)=\frac{\chi_0}{1+\alpha_a c}.
\end{equation}
We numerically solve the same problem as \cite{anderson1998continuous}, 
with the exception that 1\% amplitude red noise was added to 
the initial $c$ and $f$ fields in order to provide additional variability 
for illustrative purposes.
Further details of the numerical solutions are 
provided in the Appendix.

Figure \ref{fig:tumor} shows the dominant balances 
of the tumor angiogenesis problem of \cite{anderson1998continuous} 
at non-dimensional time $t=0.91$. The initial 
conditions are that of \cite{anderson1998continuous} 
with 1\% amplitude noise added to the initial 
angiogenic factor concentration $c$ and 
fibronectin concentration $f$ fields. $K-$means 
and CHS algorithms were employed for 
the partitioning and hypothesis selection tasks of 
the framework shown in Figure \ref{fig:algorithm}.
The optimal global 
magnitude score is $\mathcal{M}(\mathbf{E},\mathbf{H})=0.96$ 
and it occurs for $K=9$. The dominant balances indicate that 
the regions of 
fastest growth/decay of endothelial cell concentration $n$
occur at the front at approximately $x=0.4$, where the 
$\chi n \nabla^2c$ and $\rho_a\nabla n \cdot \nabla f$ terms 
are inactive (cluster 4, purple). 
The dominant balances also indicate that in 
regions of weaker growth/decay, the time rate of change
endothelial cell concentration is a residual of 
a balance between $\chi\nabla n \cdot \nabla c\sim\rho_a n \nabla^2f$, 
cluster 0 (red). Figure \ref{fig:tumor} shows that the 
unsupervised learning framework proposed in this Article 
provides a more nuanced and direct method for 
analyzing the effect of individual terms on the growth of a 
specific variable than running a multitude of simulations with 
different terms eliminated, as was done by \cite{anderson1998continuous}.

\subsubsection{Spatially-developing turbulent boundary layers}
\begin{figure}[h!]
\centering
\includegraphics[width=1\textwidth]{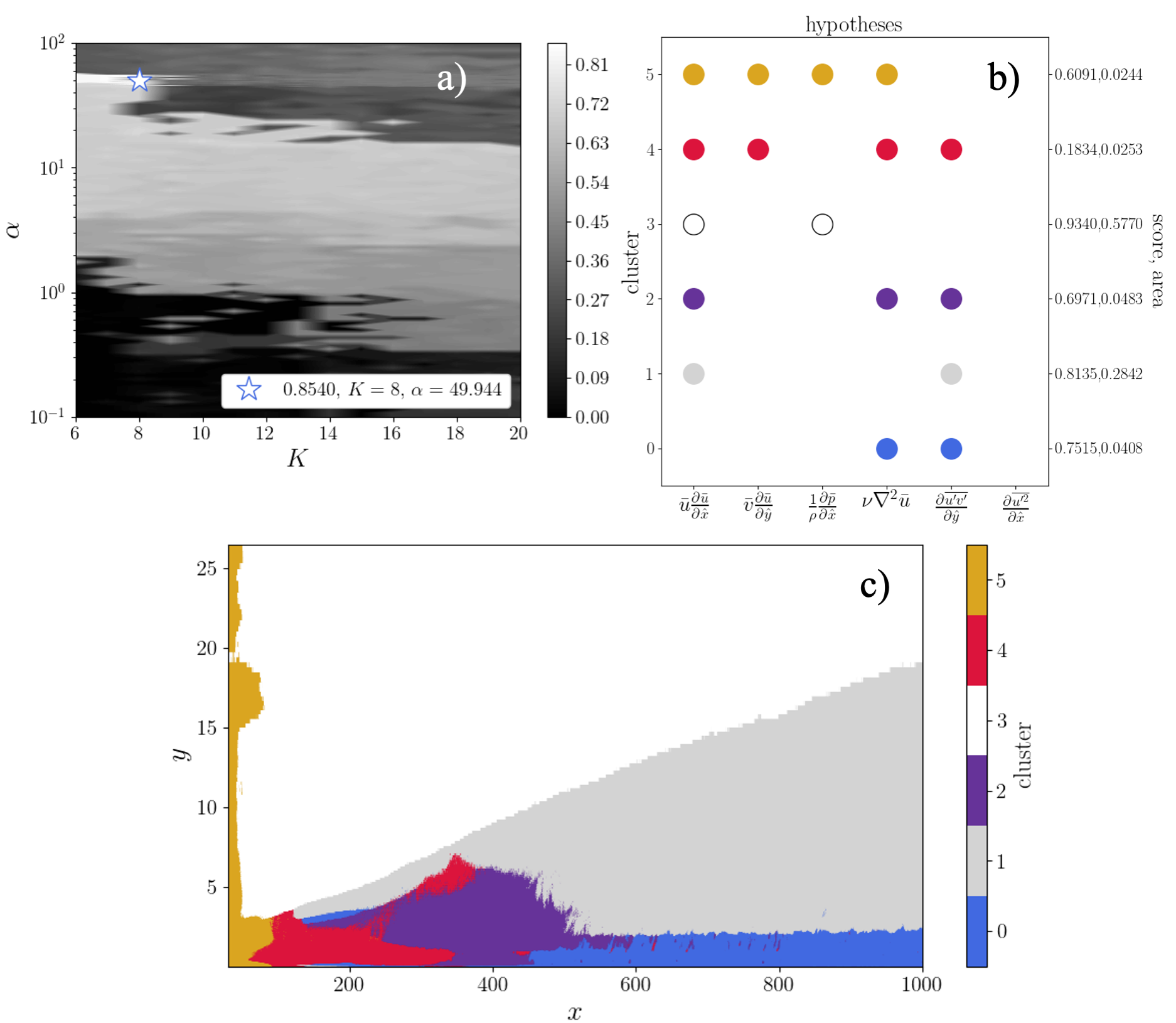}
\caption{The dominant balances of the spatially-developing turbulent boundary layer. 
Figure $a)$ shows the global score $\mathcal{M}(\mathbf{E},\mathbf{H})$ as the prescribed number of components for GMM clustering, $K$, and $l_1$ LASSO regression coefficient for SPCA, $\alpha$, are varied. The optimal result is $\mathcal{M}(\mathbf{E},\mathbf{H})=0.85$ for $K=8$ and $\alpha=49.94$. 
Figures $b)$ and $c)$ show that the optimal is consistent with the results shown in \cite{callaham2021learning} and with domain knowledge, where the blue cluster is the viscous sublayer, 
the grey cluster is the region of the so-called law of the wall, and the white cluster is the free stream flow.   
}
\label{fig:tbl}
\end{figure}
In this section we apply the framework in Figure 
\ref{fig:algorithm} to one of the dynamical regime problems 
that 
\cite{callaham2021learning} solved by using 
Gaussian Mixture Model (GMM) clustering and SPCA: the 
turbulent boundary layer (TBL) that develops as a high speed 
flow blows over a flat plate.
We use the same data set as \cite{callaham2021learning}:
the turbulent boundary layer direct numerical simulation data available in the 
Johns Hopkins University turbulence database (\cite{zaki2013streaks}).
The equation that governs the velocity 
in the direction of the mean flow, $u$, also known as the boundary layer equation  (\cite{tennekes1972first}), is
\begin{equation}
    \overbrace{\overline{u} \frac{\partial \overline{u} }{\partial \hat{x}} +
{ \overline{v} \frac{\partial \overline{u} }{\partial \hat{y}} }}^{ \substack{\text{mean}\\\text{momentum flux}\\\text{divergence}}}\hspace{2mm}
=\hspace{-4mm}\overbrace{- \frac{1}{\rho}\frac{\partial \overline{p} }{\partial \hat{x}}}^{ \substack{\text{mean}\\\text{pressure gradient}}}
\hspace{-2.5mm}+\overbrace{\nu\nabla^2\overline{u}}^{ \substack{\text{mean}\\\text{momentum}\\\text{diffusion}}}
-\hspace{1mm}
\overbrace{\frac{\partial \overline{u'v'} }{\partial \hat{y}} 
-\frac{\partial \overline{u'^2} }{\partial \hat{x}}}^{ \substack{\text{turbulent}\\\text{momentum flux}\\\text{divergence}}}, 
\label{eq:tbl}
\end{equation}
where the velocity and pressure fields ($u,v,p$) have been 
decomposed into mean and fluctuating components denoted by 
overbars and primes, respectively: the $x$ direction points in 
the downwind direction, and the $y$ direction points in the direction normal
to the surface. The averaging operator represents averaging over 
the spanwise direction as well as averaging over time, and 
the diffusion operator is defined as $\nabla^2=\partial^2/\partial \hat{x}^2+\partial^2/\partial \hat{y}^2$. $\rho$ and $\nu$ are constants 
that represent the fluid density and kinematic viscosity, respectively. 

The optimal results shown in Figure \ref{fig:tbl} are qualitatively consistent with 
and quantitatively similar to 
the results of \cite{callaham2021learning}, with the notable difference that here the optimal clustering and dimensionality reduction parameters
are defined by the objective function instead of by the user's knowledge of the algorithms. 
Clusters 0, 1, and 3 (blue, grey, and white, respectively) are consistent with 
established domain knowledge of viscous sublayers, the law of the wall, 
and free stream flow, respectively (\cite{schetz2011boundary})
but have been identified by a completely objective technique for the first time. Slight 
differences between the results shown in Figure \ref{fig:tbl} and \cite{callaham2021learning} arise from standardization of the equation data prior to clustering and 
from normalization of the cluster data prior to dimensionality reduction.

\section{Time complexity}
While our framework (Figure \ref{fig:algorithm}) and proposed objective function (Equation \ref{eq:global_score}) yield robust solutions to Equation \ref{eq:problem_formulation} for data sets with obvious 
dynamical regimes 
(e.g. Section \ref{sec:synth_data}), the time complexity of the framework 
depends upon the chosen clustering algorithm and 
hypothesis selection  
algorithms.
A comprehensive analysis of all 
possible algorithm choices is beyond the scope of this 
Article, though we can infer general 
properties of 
the complexity of 
the framework. 
The search over hyperparameters may very well 
be NP-hard; indeed, the search over hyperparameter $K$, 
the prescribed 
number of clusters for $K-$means, 
to minimize the sum of the square of the Euclidean distance of each data point to its nearest center
is NP-hard even for just two features, $D=2$
(\cite{mahajan2012planar}).
Furthermore, a n{\"a}ive search 
over a potentially 
infinite number of hyperparameters is obviously unfeasible, therefore
practical implementation
of the framework 
depends upon 
the familiarity of the user with 
the chosen algorithms and the statistical properties 
of the given data set.

 \begin{figure}[h!]
 \includegraphics[width=1.\textwidth]{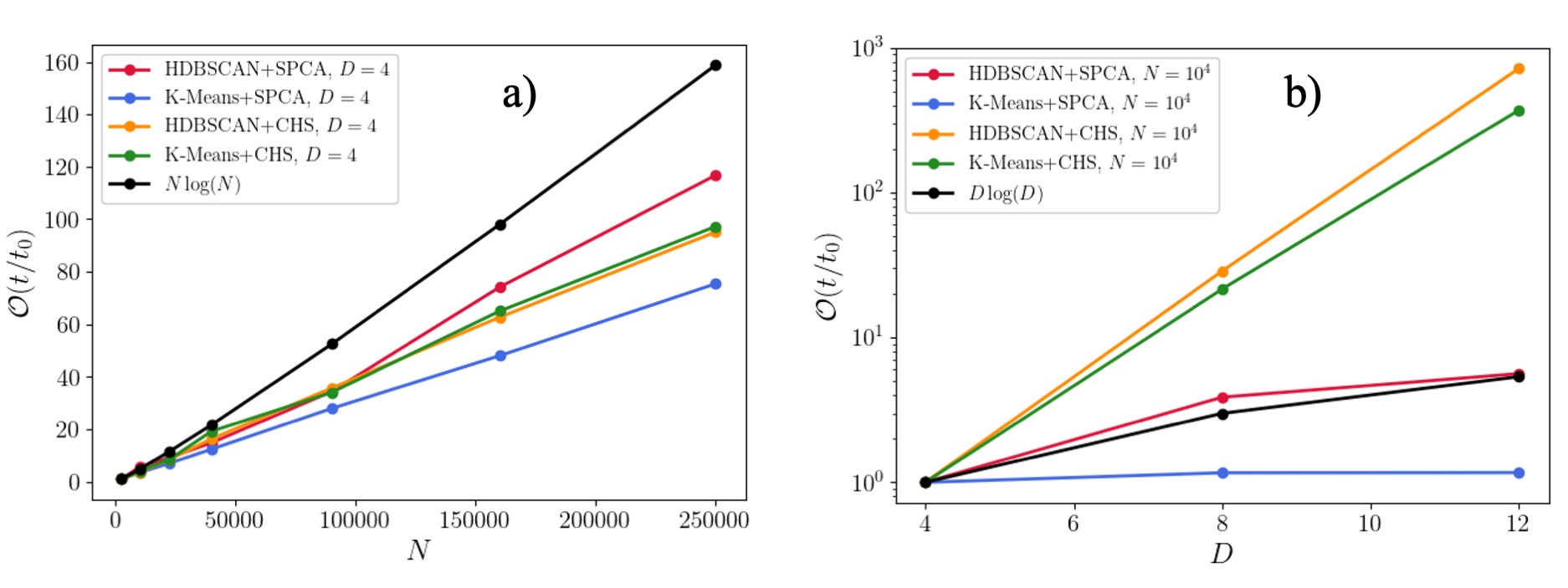}
 \caption{Time complexities of the framework 
 shown in Figure \ref{fig:algorithm} for 
 different clustering and hypothesis selection 
 algorithms as a function of the size of the input data set $N$ ($a$) and as a function of the 
 size of the number of equation terms $D$ ($b$) 
 when computed using the synthetic data described 
 in Equations \ref{eq:synth1}-\ref{eq:synth3}
and Table \ref{table:1}. The optimal dominant balances, their distributions, and global magnitude score for all complexity examples are identical.} 
 \label{fig:complexity}
 \end{figure}

The time complexity of different clustering 
and hypothesis selection algorithm choices 
applied to the synthetic data problem 
(Figure \ref{fig:synth}) is shown in Figure \ref{fig:complexity}. 
Each data point in Figures \ref{fig:complexity}a) 
and \ref{fig:complexity}b) represents the 
average wall clock time for one pass through 
the framework in Figure \ref{fig:algorithm}, 
i.e. for a single set of hyperparameters.
Figure \ref{fig:complexity}$a)$ shows that combining SPCA (\cite{zou2006sparse} with either
$K-$means by Expectation-Maximization (\cite{bishop2006pattern}) or HDBSCAN (\cite{campello2015hierarchical}), which are parametric and non-parametric clustering algorithms, respectively, results in computational times that scale polynomially in the data set sample size $N$.
In Figures \ref{fig:complexity}$a)$ and \ref{fig:complexity}$b)$, the computational time spent on one cycle for a given $N$ or $D$, by each algorithmic choice, is normalized by $t_0$, the computational time taken for the smallest $N$ or $D$, respectively.

As expected by its construction, Figure \ref{fig:complexity}$b)$ shows 
that the CHS is prohibitively complex as the number of equation terms increases because its complexity 
scales with $\mathcal{O}(2^D)$. However, 
the advantage of 
the CHS algorithm is that it has no hyperparameters; therefore, for equations  with only a handful of terms, between 3 to 5 terms, we recommend using CHS over SPCA, which has a LASSO regression coefficient that must be optimized in addition to clustering algorithm parameters. For equations with more terms, we recommend SPCA, but there are many other possible choices, viz. \cite{van2009dimensionality}. 

In summary, our framework yields robust solutions to Equation \ref{eq:problem_formulation} for data sets with obvious 
dynamical regimes 
(e.g. Section \ref{sec:synth_data}), but 
efficient computation requires the user to consider the size of 
the data set $N$ and the number of equation terms $D$ when selecting 
clustering and hypothesis selection algorithms.

\section{Conclusions}
We have formalized the problem of identifying dynamical regimes 
as an optimization problem in which 
data, consisting of the terms in an equation that describes the complete dynamics of a physical process,
is partitioned into clusters of unique, reduced sets of terms that are dominant in the equation. 
The optimization 
is achieved by maximizing an objective function that we have defined such
that it favors large magnitude 
differences between selected and neglected terms, 
and penalizes large magnitude spread within the 
selected terms. Our formulation of the problem 
and our definition of the objective function are independent 
of the method chosen to label each observation 
as a dominant balance and, together, they transform 
a form of analysis of data of dynamical systems that was previously an \textit{ad hoc}, heuristic notion, into an objective method for the 
identification 
of non-asymptotic dynamical regimes.

We demonstrated how this formulation of the 
problem can be applied to an unsupervised learning 
framework, in which equation data is partitioned 
by clustering algorithms (\cite{sonnewald2019unsupervised}, \cite{callaham2021learning}), 
labeled by hypotheses generated by a dimensionality 
reduction algorithm (\cite{callaham2021learning}), 
and finally optimized over algorithm parameters 
using our objective function. This was demonstrated on 
synthetic data with obvious dominant terms separated by 
a sharp regime boundary as well as with three examples 
drawn from climate dynamics, cancer modeling, and 
aerodynamics. We demonstrate 
how the dynamical regime problem can be solved 
by an unsupervised learning framework 
for fixed algorithm parameters, and we note that a
n{\"a}ive search of possible infinite algorithm parameter space 
for a global optimum is computationally unfeasible. 
We have effectively exchanged 
prerequisite domain knowledge with 
a familiarity with and skill at using
clustering and dimensionality reduction algorithms, thus 
opening the door to rapid dynamical regime discovery within 
large data sets.

\newpage 
\section{Acknowledgements}
This work was performed under the auspices of DOE. Financial support comes partly from Los Alamos National Laboratory (LANL), Laboratory Directed Research and Development (LDRD) project "Machine Learning for Turbulence," 20180059DR. LANL, an affirmative action/equal opportunity employer, is managed by Triad National Security, LLC, for the National Nuclear Security Administration of the U.S. Department of Energy under contract 89233218CNA000001. Computational resources were provided by the Institutional Computing (IC) program at LANL. 

MS acknowledges funding from Cooperative Institute for Modeling the Earth System, Princeton University, under Award NA18OAR4320123 from the National Oceanic and Atmospheric Administration, U.S. Department of Commerce.
The statements, findings, conclusions, and recommendations are those of the authors and do not necessarily reflect the views of Princeton University, the National Oceanic and Atmospheric Administration, or the U.S. Department of Commerce.

\section{Appendices}
\subsection{Parameter ranges for time complexity calculations}
The average wall time elapsed for the framework computations over hyperparameter ranges are shown in Figure \ref{fig:complexity}. The hyperparameter ranges were specfied as follows. 
For $K-$means clustering the number of prescribed clusters $K$ was specified as $K=\{2,...,10\}$ and the other hyperparameters were the default choices as provided by SciKit Learn (\cite{scikit-learn}).
For HDBSCAN clustering the prescribed minimum 
number of samples for a cluster was specified as 
100 samples, and the minimum cluster size was 
varying from 2000 samples to 3000 samples. 
For hypothesis selection by SPCA, the LASSO 
regression coefficient was varied between $10^{-2}$ and $10^2$.

\subsection{Tumor-induced angiogenesis simulation}
A second-order accurate finite difference code was used to 
calculate each term in the expanded form of the endothelial cell density 
equation, such that $\mathbf{E}$ is composed of observations 
of the terms in 
Equation (\ref{eq:expanded_EC_density}). We employ the same 
boundary conditions, initial conditions, and
constant coefficients ($D_a$, $\alpha_a$, $\chi_0$, $\rho_a$, $\beta_f$, $\gamma$, and $\eta_c$) as \cite{anderson1998continuous} at double 
the resolution. 
Second-order finite differences were employed for spatial derivatives and 
4th-order adaptive 
Runge-Kutta was employed for the temporal evolution.
No-flux boundary conditions were applied to all four boundaries of the square domain:
\begin{equation}
    \mathbf{n}\cdot(D_a\nabla n
  -\chi(c)n\nabla c
  -\rho_a n\nabla f)=0,
\end{equation}
where $\mathbf{n}$ is the unit normal vector to the boundaries.
The initial conditions, for a circular tumor 
some distance 
from three clusters of endothelial cells, are:
\begin{equation}
    c(x,y,0)=\begin{cases}
               1, \hspace{8mm} \qquad 0\leq r \leq 0.1\\
               \frac{(\nu-r)^2}{\nu-r_0}, \qquad 0.1< r \leq 1
            \end{cases},
\end{equation}
where $r=\sqrt{(x-x_0)^2+(y-y_0)^2}$.
\begin{equation}
    f(x,y,0)=k\mathrm{e}^{-\frac{x^2}{\epsilon_1}},
\end{equation}
\begin{equation}
    n(x,y,0)=\mathrm{e}^{-\frac{x^2}{\epsilon_2}}\sin^2(6\pi y),
\end{equation}
where $\nu=(\sqrt{5}-0.1)/(\sqrt{5}-1)$, $r_0=0.1$, $x_0=1$, $y_0=1/2$, $k=0.75$, $\epsilon_1=0.45$, $\epsilon_2=0.001$.
The constant coefficients were specified as $D_a=0.00035$, $\alpha_a=0.6$, $\chi_0=0.38$, $\rho_a=0.34$, $\beta_f=0.05$, $\gamma=0.1$, and $\eta_c=0.1$.

\newpage
\bibliographystyle{apalike}   
\bibliography{ref}  
\end{document}